\documentclass[sigconf]{acmart}
\AtBeginDocument{%
  }

\usepackage{acronym}
\usepackage{pbalance}

\acrodef{TAM2}{Technology Acceptance Model Version 2}
\acrodef{ML}{Machine Learning}
\acrodef{STT}{Speech-to-Text}
\acrodef{TTS}{Text-to-Speech}
\acrodef{ASR}{Automatic Speech Recognition}
\acrodef{RAG}{Retrieval Augmented Generation}
\acrodef{HRI}{Human-Robot Interaction}
\acrodef{ITU}{intention to use}
\acrodef{PU}{perceived usefulness}
\acrodef{PEOU}{perceived ease of use}
\acrodef{UT}{UsefulnessToday}

\copyrightyear{2026}
\acmYear{2026}
\setcopyright{cc}
\setcctype{by-nc-nd}
\acmConference[IVA 2026]{ACM International Conference on Intelligent Virtual Agents}{September 07--11, 2026}{Puebla, Mexico}
\acmBooktitle{ACM International Conference on Intelligent Virtual Agents (IVA 2026), September 07--11, 2026, Puebla, Mexico}
\acmDOI{10.1145/3806774.3827982}
\acmISBN{979-8-4007-2647-7/2026/09}

\begin{document}

\title[Back to the museum]{Back to the museum: Investigation of the acceptance of Android Andrea with and without emotion simulation in a museum}

\author{M. Heisler}
\authornotemark[1]
\email{heisler@hdm-stuttgart.de}
\orcid{0009-0004-7982-1000}
\author{C. Becker-Asano}
\authornotemark[1]
\email{becker-asano@hdm-stuttgart.de}
\orcid{0000-0002-9946-0458}
\affiliation{%
 \institution{Hochschule der Medien}
 \city{Stuttgart}
 \country{Germany}
}

\renewcommand{\shortauthors}{Heisler et al.}

\begin{abstract}
For a second time, the android robot Andrea was set up at a public museum in Germany for six consecutive days to have conversations with visitors, fully autonomously. Building on previously gathered qualitative results, the robot was now capable of engaging in multi-lingual conversation with the visitors about the museum context. The robot was prepared with context information about the museum in general and its surrounding exhibits this time. The robot featured a slightly artificial sounding voice that was previously evaluated as congruent with its gender-ambiguous but very humanlike design. Three experimental conditions were implemented, in which either (1) the robot simulated no emotions, (2) the robots emotions were determined by ChatGPT 4.1, or (3) the WASABI emotion simulation architecture simulated the robot's emotion dynamics. An extended version of the \textit{\ac{TAM2}} questionnaire was employed to let 73 visitors report on several factors of their opinion about the android robot Andrea after having experienced it. In result, the statistical analysis suggests that these first two approaches to implementing emotions into the chat architecture of our android robot Andrea did not yield any positive effects on the subjective evaluations by the visitors and were not detectable on a conscious level. 
\end{abstract}

\ccsdesc[500]{Human-centered computing~Empirical studies in HCI}
\ccsdesc[300]{Human-centered computing~Empirical studies in interaction design}

\keywords{Human-Robot Interaction, empirical study, android robot, technology acceptance model}
\begin{teaserfigure}
  \includegraphics[width=\textwidth]{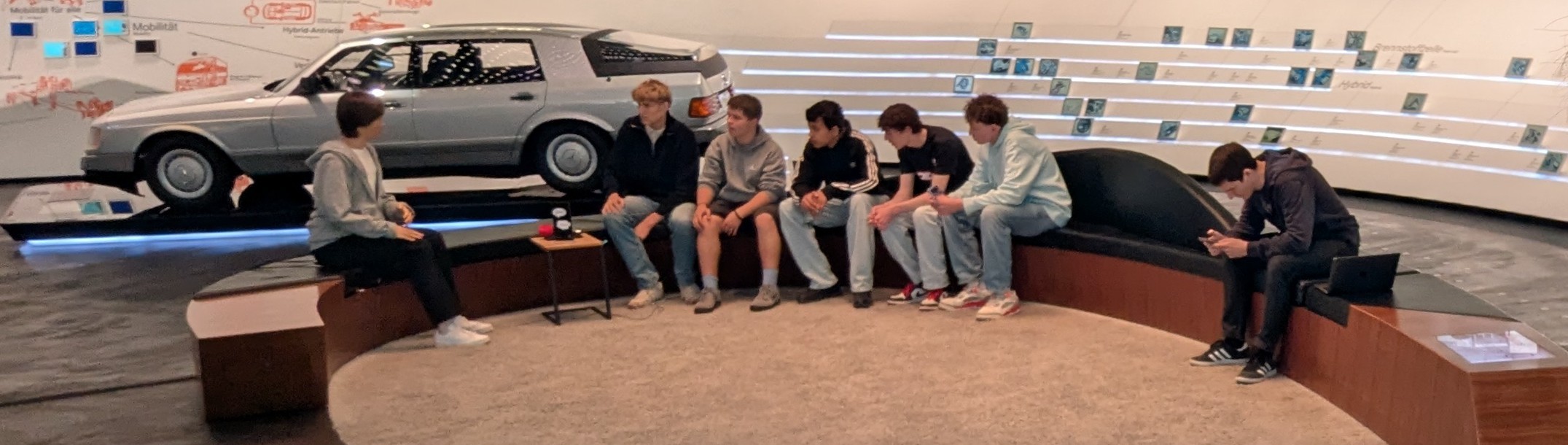}
  \caption{Android Andrea sitting on a bench in the museum and conversing with visitors}
  \Description{An android robot sitting on a bench inside a museum and interacting with a group of young people.}
  \label{fig:teaser}
\end{teaserfigure}

\maketitle

\section{Introduction}
The intentional anthropomorphic design of android robots is meant to facilitate sophisticated multimodal communication. Consequently, they are presently being tested in scenarios where smooth, human‑like interaction is crucial—for instance, to offer companionship and information to elderly users \cite{cavallo_not_2022, thalmann_nadine_2021} or to serve in roles such as interviewers and receptionists \cite{kawahara_intelligent_2021, baka_social_2022}.

In 2023, the android robot Andrea operated autonomously for six days at a German museum, engaging 44 visitors in unstructured conversations that later fed into structured interviews \cite{heisler_conversations_2025}. Visitors most frequently sought exhibit information, and interviewees highlighted faster response times and multilingual support as key improvements; gender‑changing cues had little effect on overall perception. 

Building upon these previous results, this paper presents a number of technical improvements regarding the software architecture that drives the multimodal and multi-lingual behaviour of the android robot. This includes the integration of two different approaches to simulate emotions that are expressed by the robot. In the aim to evaluate the impact of the emotion simulation on the visitors' perception, the results derived from 73 questionnaires are being reported here.

The remainder of this paper is structured as follows. In Section~\ref{sec:related} work that is related is being discussed. Section~\ref{sec:technical} explains how the android robot was set up to answer visitor questions autonomously, before in Section~\ref{sec:method} details of the questionnaire are presented. The survey results are provided in Section~\ref{sec:survey} and discussed in Section~\ref{sec:discussion}.

\section{Related work}
\label{sec:related}

Before the availability of large language models such as ChatGPT, natural language processing for interactive, autonomous, social robots has been a major challenge \cite{hellou_technical_2022}. Most early social robots did not feature a very human-like design, but a mobile platform enabling them to guide visitors through the museum.  

Android robots are a specialized subset of the broader category of social robots, many of which have been deployed in museum settings. A review of social robots in museums \cite{gasteiger_deploying_2021} identified that the predominant role of these machines is to act as guides; they are less frequently used for entertainment or educational purposes. The study also found that the majority of such robots are deemed acceptable for museum environments and distilled three key themes from the literature that underpin successful human‑robot interaction in these contexts, namely appropriate facial expressions, movement, and communication abilities. 


In Linz, Austria, the android Geminoid HI‑1 was positioned behind a table in the public café atop the Ars Electronica museum, accompanied by informational panels about Japan \cite{rosenthal2014uncanny}. Three operating modes were tested: (1) passively monitoring a laptop placed in front of it, (2) autonomously making eye contact with passersby, and (3) being tele‑operated. Structured interviews and video analyses revealed that increased eye contact made the robot more readily identifiable as a machine, while participants tended to find it more interesting than unsettling. 

Later, Geminoid HI‑1 was tele‑operated during the Ars Electronica festival to engage with visitors \cite{becker-asano_exploring_2010}. Qualitative interviews indicated a predominance of positive over negative remarks about the robot; 37.5\% of respondents reported feeling uncanny, whereas 29\% enjoyed conversing with it.

To probe the Uncanny Valley phenomenon, museum patrons interacted with a tele‑operated Telenoid robot that had been installed in the Ars Electronica Center’s public robot laboratory in 2015 \cite{mara_science_2015}. Unlike most androids, this robot’s look is far less human‑like. The study found that when visitors were first given a science‑fiction narrative framing the encounter, the robot’s eeriness ratings dropped noticeably.

\textit{Nadine the Social Robot} was housed at Singapore’s ArtScience Museum in 2017 for a five‑month period, but no formal user study could be conducted because of a confidentiality clause \cite{thalmann_nadine_2021}. The same android robot was recently tested in a semi-controlled setup inside the lobby of the University of Geneva \cite{kang_affective_2026}. The questionnaire results suggest that an LLM‑powered hyper‑realistic robot enhances users’ perceptions of pleasantness and approachability while slightly reducing creepiness, with conversational naturalness and interest‑keeping being the primary drivers of engagement. It recommends prioritizing fluid, engaging dialogue over mere human‑like mimicry and calls for more diverse, objective, and long‑term studies to confirm these findings.

Research on social humanoid robots such as Ameca \cite{berns_you_2024} demonstrates that the uncanny valley effect depends on a delicate interplay between appearance, emotional expression, and interaction skills. These findings suggest that balancing human‑like features with functional behavior and adaptive emotion can improve usability and mitigate uncanny feelings.
Ameca robots are also exhibited in various museums, like Heinz Nixdorf MuseumsForum\footnote{\url{https://www.hnf.de/dauerausstellung/ausstellungsbereiche/global-digital/mensch-roboter-leben-mit-kuenstlicher-intelligenz-und-robotik/ameca.html}}, Zukunftsmuseum Nürnberg\footnote{\url{https://www.deutsches-museum.de/nuernberg/aktuell/roboter-ameca-im-zukunftsmuseum}} or Museum of the Future, Dubai\footnote{\url{https://gulfnews.com/uae/science/dubai-museum-of-the-futures-upgraded-ai-powered-humanoid-robot-speaks-hindi-too-1.500104771}}, but no formal user study evaluating the presence of the robot was conducted.

In 2023, an autonomous run of six days saw the android Andrea deployed in a German museum, where it conversed freely with 44 visitors \cite{heisler_conversations_2025}. The dialogue data revealed that most guests used Andrea primarily to inquire about exhibit details, and the analysis results of structured interviews suggest that visitors find quicker replies and multilingual capability to be necessary upgrades. Notably, variations in gender‑changing cues did not markedly influence how the robot was perceived.

This previous work shows the need to research the effectiveness and acceptance of more emotionally interactive, android robots employed in public spaces such as a museum.

\section{Technical setup}
\label{sec:technical}

\begin{figure}
    \centering
    \includegraphics[width=\linewidth]{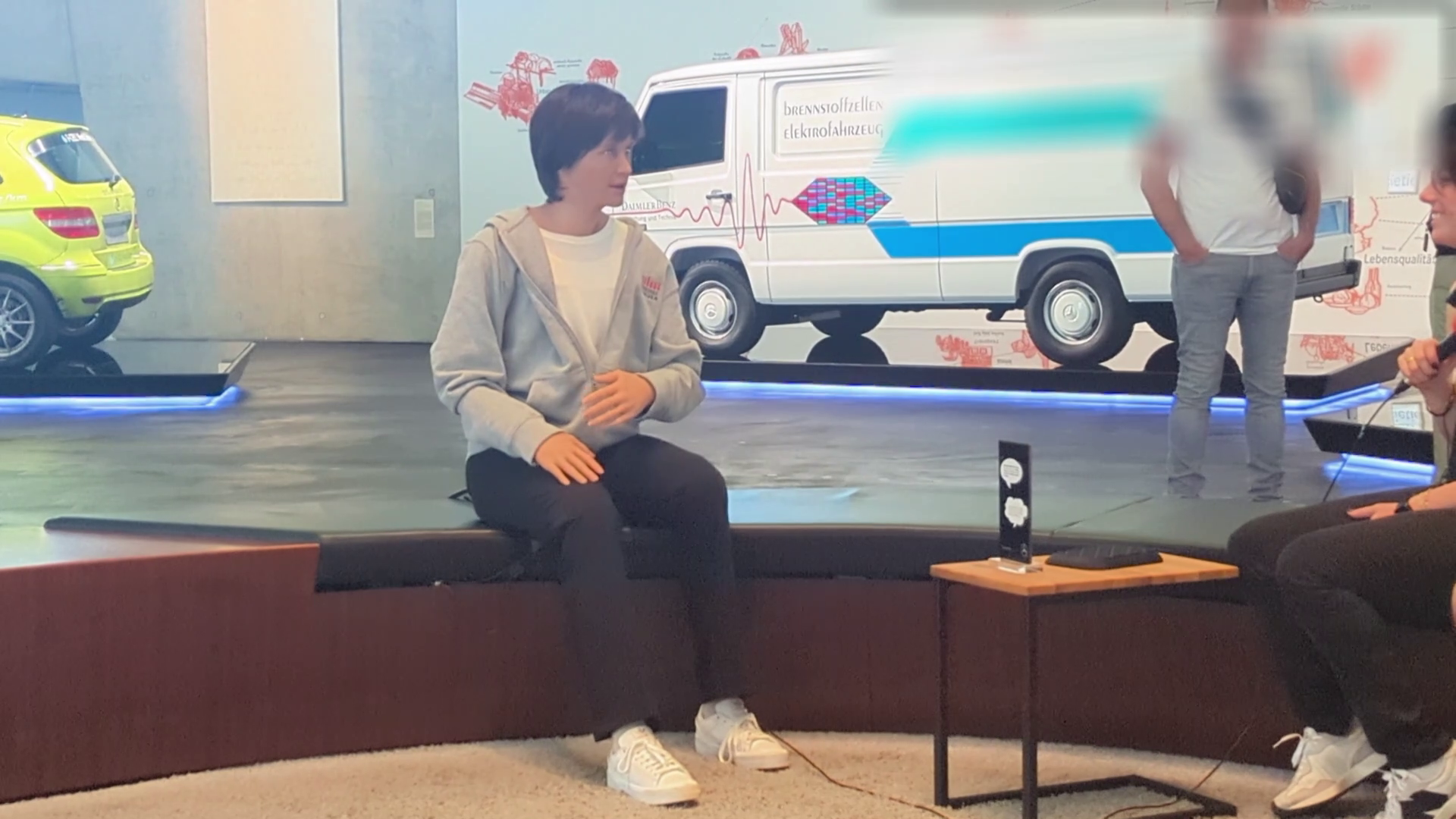}
    \caption{The android robot placed on a bench in the museum with a microphone for verbal interaction}
    \Description{The android robot is sitting on a bench inside the museum. Cars are behind the robot. In the far right of the image a person is holding the microphone to talk to the robot. In front the robot is a small table to place the microphone when not interacting.}
    \label{fig:Andrea}
\end{figure}

\begin{figure*}[ht]
    \centering
    \includegraphics[width=0.8\linewidth]{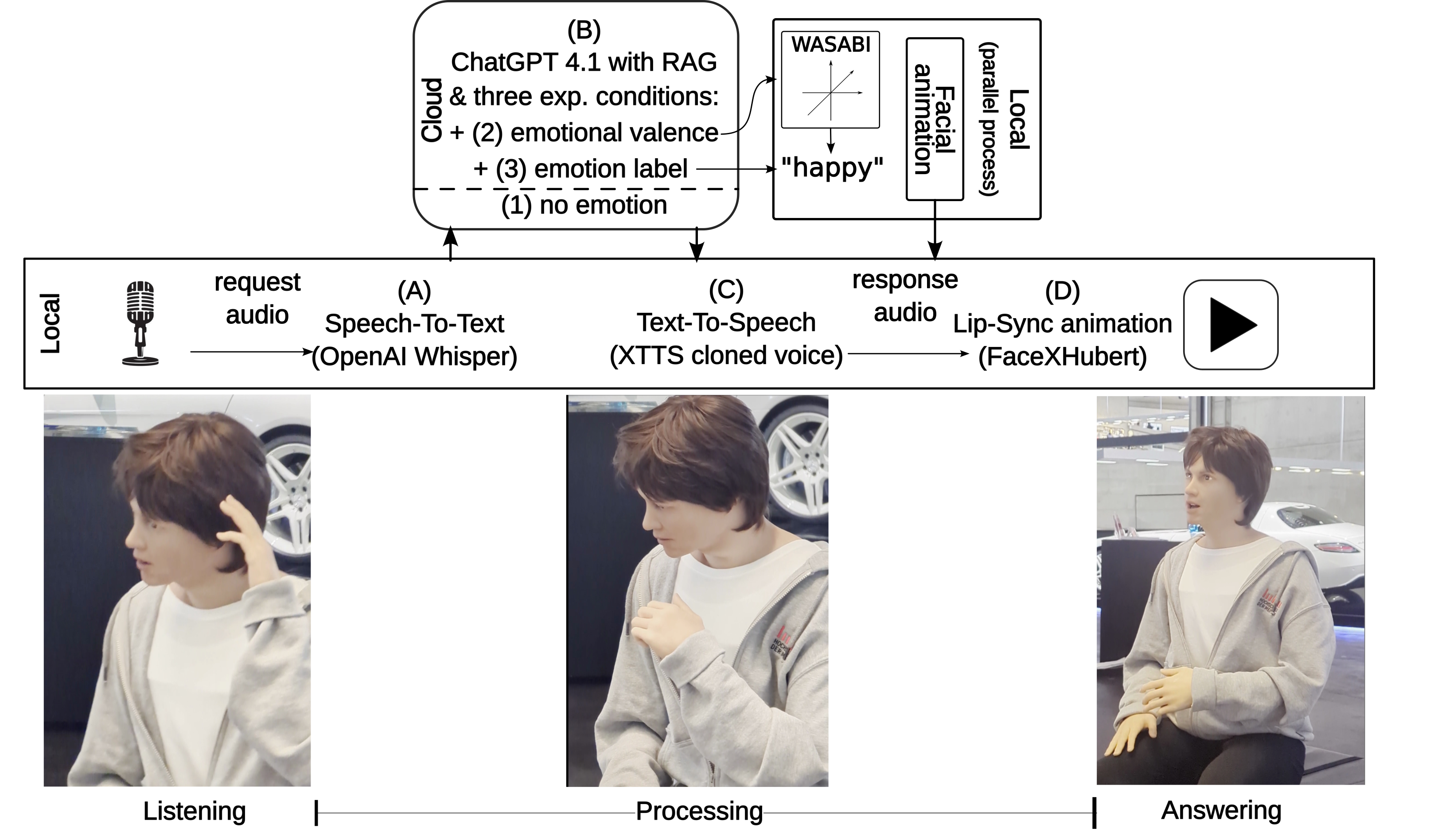}
    \caption{Overview of the software architecture of the conversational functions of the android Andrea.}
    \Description{The software architecture consisting of a pipeline of STT, ChatGPT, TTS and Lip-sync animation is depicted as described in the text. Also the emotion handling using WASABI or labels directly predicted by ChatGPT and mapped to facial expressions is included in the depiction. Example images of the robot while listening (left arm at left ear), processing (fist below chin and looking side-ways downward) and answering (arms lowered, face and posture upright and mouth open) are shown below the corresponding steps in the pipeline.}
    \label{fig:Andrea_swarchitecture}
\end{figure*}

A very human-like robot was installed in the Mercedes Benz Museum in Stuttgart, Germany, for six consecutive days freely accessible for the visitors. The android robot was seated on a bench in one of the main exhibition halls of the museum, called Mythos 6, with experimental cars of the museum owner's brand surrounding it.

The android robot was assembled by the Japanese firm A‑Lab. Its sitting posture is comparable to the company’s Premium Model, but it features additional actuators for independent control of each eyelid and finger. Altogether, 52 pneumatic actuators manage the upper body and facial movements. 

The robot’s exterior is deliberately androgynous, and a similarly androgynous voice was designed for it in previous research \cite{kuch_evaluating_2024}. It wears casual blue jeans and a gray zip‑up hoodie emblazoned with a small university logo (see Fig.~\ref{fig:teaser} and \ref{fig:Andrea}). 

Cameras are mounted in both of the robot's eyes to capture visual input. Meanwhile, conversational interaction is facilitated by a hidden speaker concealed inside the hoodie and a microphone placed on a small table directly in front of the android. The onboard software runs on a single Nvidia Jetson Orin housed in the bench under the robot. Power is supplied by electricity and compressed air (7 bar), while data communication with the Jetson is established via a USB connection. The compressor is placed inside a custom-build, black counter desk that is placed approximately five meters behind the back of the robot and is covered with convoluted acoustic foam inside to dampen the noise.

The software components are an extended version of \cite{heisler_conversations_2025, heisler_android_2023}, which combines four separate, open-source \ac{ML} models and OpenAI's ChatGPT assistant function, see also Fig.~\ref{fig:Andrea_swarchitecture}:
\begin{enumerate}
    \item[(A)] For \ac{STT} \verb|whisper-large-v3-turbo| is used supporting multiple languages as input. The input text is then transmitted to an OpenAI assistant.
    \item[(B)] The user request is taken as input for a ChatGPT4.1-based OpenAI-assistant that is equipped with a special system prompt and uses \ac{RAG} as described below. Also, the three different experimental conditions are realized in this step.
    \item[(C)] Multilingual output is generated using XTTSv2 \cite{casanova24_interspeech} for speech synthesis. This \ac{ML} model supports 17 languages. A specific design-congruent voice \cite{kuch_your_2025} was cloned and used consistently in all languages.
    \item[(D)] The generation of lip-sync facial animation is described in \cite{heisler_making_2023}. To speed up this process, the \ac{ML} model used previously is replaced by FaceXHubert \cite{haque_facexhubert_2023}.
    \item[(E)] The people tracking is based on a realtime analysis of the video stream provided by the camera in the robot's left eye using \verb|posenet| \cite{kendall_posenet_2015} (not shown in Fig.~\ref{fig:Andrea_swarchitecture}).
\end{enumerate}

For initiating a dialogue the user must manually enable the microphone, which triggers a listening animation. We deliberately avoided passive listening to reduce the likelihood of spurious activations in noisy public spaces. Once the microphone is disabled, the robot switches to a “thinking” animation, signaling that it is processing the user’s input. As shown in \cite{namba_how_2024}, such thinking faces improve the naturalness of human‑robot interaction. To further humanize its behavior, the robot blinks at random intervals and yawns when no interaction has occurred for a while.

To generate context-sensitive textual responses, an OpenAI assistant was used with \verb|gpt-4.1| as the basis. The assistant's prompt contains information about the robot itself and about its surrounding, e.g., which cars are located where. In addition, \ac{RAG} is used to provide additional information, such as the transcripts of all audio guide data in English. 

When the robot's emotional states are simulated using WASABI \cite{becker-asano.2014}, first, the OpenAI assistant is prompted to calculate the valence of the last input sentence spoken by the user. The resulting value ranges from -100 to 100 and is sent to WASABI as a valenced impulse. In effect, WASABI, which is running as a concurrent process in the local computer, returns the emotion likelihood of the emotions ``happy'', ``sad'', ``angry'', ``fearful'', ``surprised'' or ``neutral'', of which the one with the highest value is used to animate the robot's face asynchronously. The internal emotion dynamic of WASABI lets the robot's emotional state automatically return to ``neutral'' after some time without any inputs. Alternatively, ChatGPT is instructed to generate one of these emotion labels directly. The intensity dynamics of these emotions is then taken care of in the main process of the software architecture by employing a linear decay function.

All emotions are expressed using validated static facial expressions \cite{kassner_comparing_2023}. The animations for thinking and speech have a higher priority than the emotional expressions for the relevant actuators. Thus, while the robot is speaking, its mouth movements synchronize with the speech signal; before or after speaking, however, the facial actuators reflect the simulated emotion (e.g.,~smiling).

\section{Method}
\label{sec:method}

Due to the circumstances inside a public museum, the study had to follow a between-groups design. On the first and fourth day the robot was in the "no emotion" condition, on the second and fifth day it was in the "ChatGPTemotion" emotional condition, and on the third and sixth day it was in the "WASABI" emotional condition. However, due to hardware issues the robot's facial expressions had degraded so much on the sixth day that the questionnaire data of that day had to be excluded from the analysis.

\subsection{Study procedure}
On each day, a team of three experimenters was present on-site in alternating shifts to supervise the study setup and ensure a safe environment. Depending on the situation and the experimenter on duty, interested visitors were at times actively approached, while at other times they initiated the interaction themselves. If necessary, the experimenters provided a brief introduction on how to interact with the android robot and explained the use of the microphone. 

After the interactions, users were invited to complete a questionnaire that contained an informed consent and items in both English and German. Additionally, participants were informed that they could cancel the survey at any time.
In total, $N=73$ visitors voluntarily provided questionnaire data, with $N_{ChatGPTemotion}=35$, $N_{ChatGPTpure}=24$ and $N_{WASABI}=14$, cf.~Table~\ref{tbl:Table_Descriptives_global}.

\begin{table}[!htbp]
\caption{Descriptives of the questionnaire results}
\label{tbl:Table_Descriptives_global}
\centering
\begin{tabular}{llrrrrrr}
\toprule
\multicolumn{4}{c}{~} & \multicolumn{2}{c}{Shapiro-Wilk} \\
\cmidrule{7-8}
~                   & Condition &     $Mean$ &   $SD$ &    $W$ &      $p$ \\
\midrule
ITU      & ChatGPTemotion          & 4.99 &   1.67 & 0.88 &   .001 \\
ITU      & ChatGPTpure             & 5.69 &   1.51 & 0.82 & <~.001 \\
ITU      & WASABI                  & 4.93 &   1.73 & 0.91 &   .172 \\
PU & ChatGPTemotion          & 4.19 &   1.57 & 0.95 &   .115 \\
PU & ChatGPTpure            & 5.21 &   1.74 & 0.87 &   .005 \\
PU & WASABI                 & 4.18 &    1.90 & 0.92 &   .230 \\
PEOU  & ChatGPTemotion          & 5.08 &    1.18 & 0.91 &   .010 \\
PEOU  & ChatGPTpure             & 5.89 &    0.83 & 0.89 &   .013 \\
PEOU  & WASABI                 & 5.32 &    1.69 & 0.83 &   .011 \\
UT      & ChatGPTemotion          & 7.53 &    2.72 & 0.84 & <~.001 \\
UT      & ChatGPTpure            & 7.88 &    2.11 & 0.87 &   .005 \\
UT      & WASABI                  & 7.71 &    2.81 & 0.79 &   .004 \\
\bottomrule
\end{tabular}
\end{table}

The \ac{TAM2} questionnaire from \cite{olbrecht2010} was used as a basis, as it provided the exact German translations of the original items \cite{venkatesh2000}. 
Our questionnaire can be divided into the following categories: general information such as date, time, and age; questions evaluating participants’ prior knowledge of the robot's availability at this place and whether it influenced their decision to visit the museum; \ac{TAM2} items measuring  \textit{\ac{ITU}, \ac{PU}}, and \textit{\ac{PEOU}}; questions regarding the emotions; as well as additional questions, regarding for how long participants had interacted with the robot, which languages they used, and any suggestions they had. Finally, participants were asked how useful they considered the robot in the museum context. 

\subsection{Measurement}

Whether participants knew about the robot's availability prior to their visit, and whether they had specifically come to see the robot was evaluated with yes/no questions.

The \ac{TAM2} items and emotion-related questions were evaluated using a 7-point Likert scale, where 1 indicated `strongly disagree' and 7 indicated `strongly agree'. 
\\ 
``Intention to Use'' was measured with two items: 
\begin{enumerate}
    \item Assuming I have access to the robot, I intend to use it.
    \item Given that I have access to the robot, I predict that I would use it.
\end{enumerate}
``Perceived Usefulness'' was measured with four items: 
\begin{enumerate}
    \item Using the robot improves my performance.
    \item Using the robot increases my productivity.
    \item Using the robot enhances my effectiveness.
    \item I find the robot to be useful.
\end{enumerate}
``Perceived ease of Use'' was measured with four items: 
\begin{enumerate}
    \item My interaction with the robot is clear and understandable.
    \item My interaction with the robot does not require a lot of my mental effort.
    \item I find the robot to be easy to use.
    \item I find it easy to get the robot to do what I want it to do.
\end{enumerate}
The Shapiro-Wilk test was used to determine the $W$ statistic and $p$-value, in order to assess whether the data is normally distributed, see Table~\ref{tbl:Table_Descriptives_global}.

The perceived emotionality of the robot was measured with three items:
\begin{enumerate}
    \item I have the impression that the robot displays emotional reactions
    \item I find the robot's emotional responses appropriate to the situation.
    \item The robot appears moody or shows emotional mood swings.
\end{enumerate}
The final question ``How useful do you think would it be to use this robot here today?'' (reported here as \ac{UT}) was measured by a distinct Likert scale ranging from 0 (`not at all') to 10 (`very much') to compare our setup with a previous result of a study with the robot in the same public museum \cite{heisler_conversations_2025}.

\section{Results}
\label{sec:survey}

Only five of all 73 interviewees knew about the presence of the robot beforehand and only one of them came explicitly to interact with it on that day. Nineteen interviewees were in the youngest age range between 10 and 20 years old (\textit{ChatGPTpure} four, \textit{ChatGPTemotion} 13, and \textit{WASABI} two), 22 were between 21 and 30 years of age (\textit{ChatGPTpure} four, \textit{ChatGPTemotion} 12, and \textit{WASABI} six), and the remaining 25 were older than 30 years. The robot was spoken to in German 788 times and in English 659 times, followed by Turkish (102), Spanish (82), and Russian (61) times. It responeded most often in German by producing a total of 2023 German sentences, followed by English (1212), Turkish (324), Spanish (194), and Russian (162). When the input sentence was very short, the language detection failed sometimes and the robot answered in a different language, though.

\begin{figure*}[ht] 
\centering
\begin{minipage}{0.48\textwidth}
    \centering
    \includegraphics[width=0.9\linewidth]{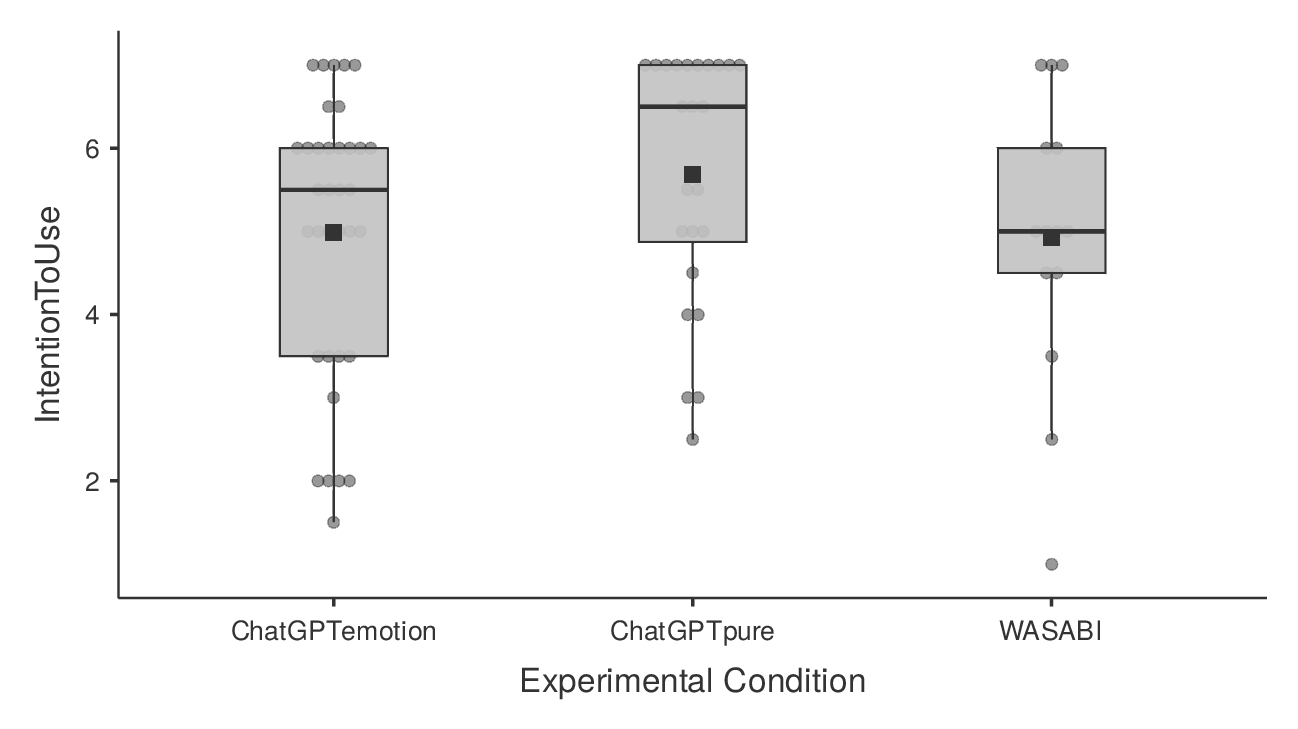}
    \caption{Box plot showing the distribution of \textit{\acf{ITU}} by experimental condition with indicated median values.}
    \Description{Box plots visualizing the descriptive data also reported in Table 1.}
    \label{intention_to_use_loc}
\end{minipage}\hfill
\begin{minipage}{0.48\textwidth}
    \centering
    \includegraphics[width=0.9\linewidth]{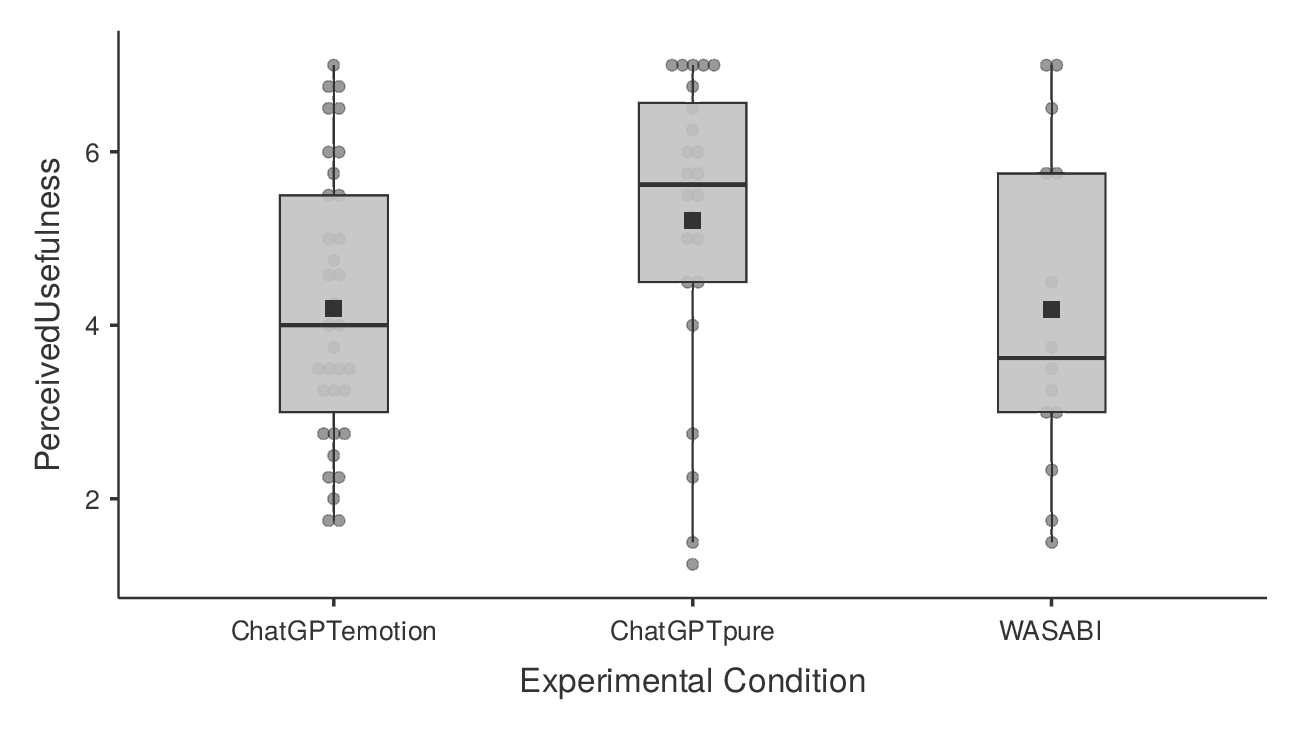}
    \caption{Box plot showing the distribution of \textit{\acf{PU}} by experimental condition with indicated median values.}
    \Description{Box plots visualizing the descriptive data also reported in Table 1.}
    \label{perceived_usefulness_loc}
\end{minipage}
\begin{minipage}{0.48\textwidth}
    \centering
    \includegraphics[width=0.9\linewidth]{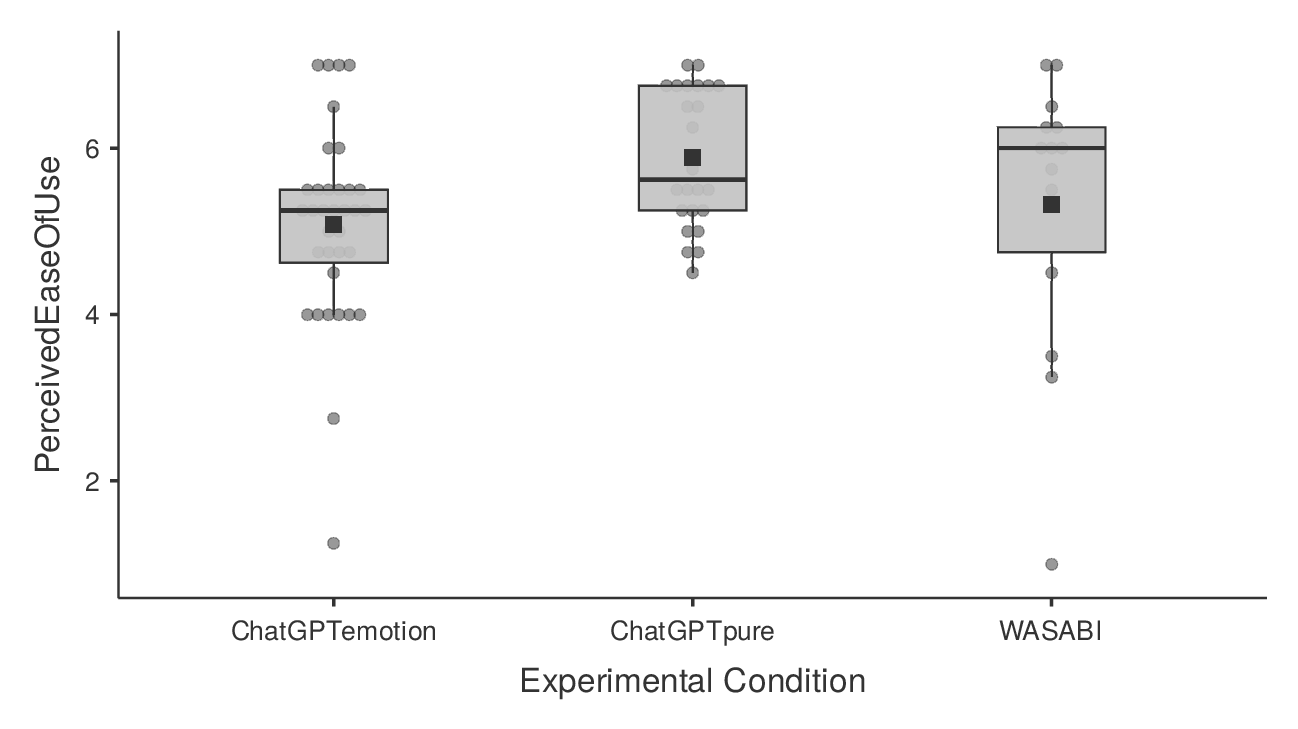}
    \caption{Box plot showing the distribution of \textit{\acf{PEOU}} by experimental condition with indicated median values.}
    \Description{Box plots visualizing the descriptive data also reported in Table 1.}
    \label{perceived_ease_of_use_loc}
\end{minipage}\hfill
\begin{minipage}{0.48\textwidth}
    \centering
    \includegraphics[width=0.9\linewidth]{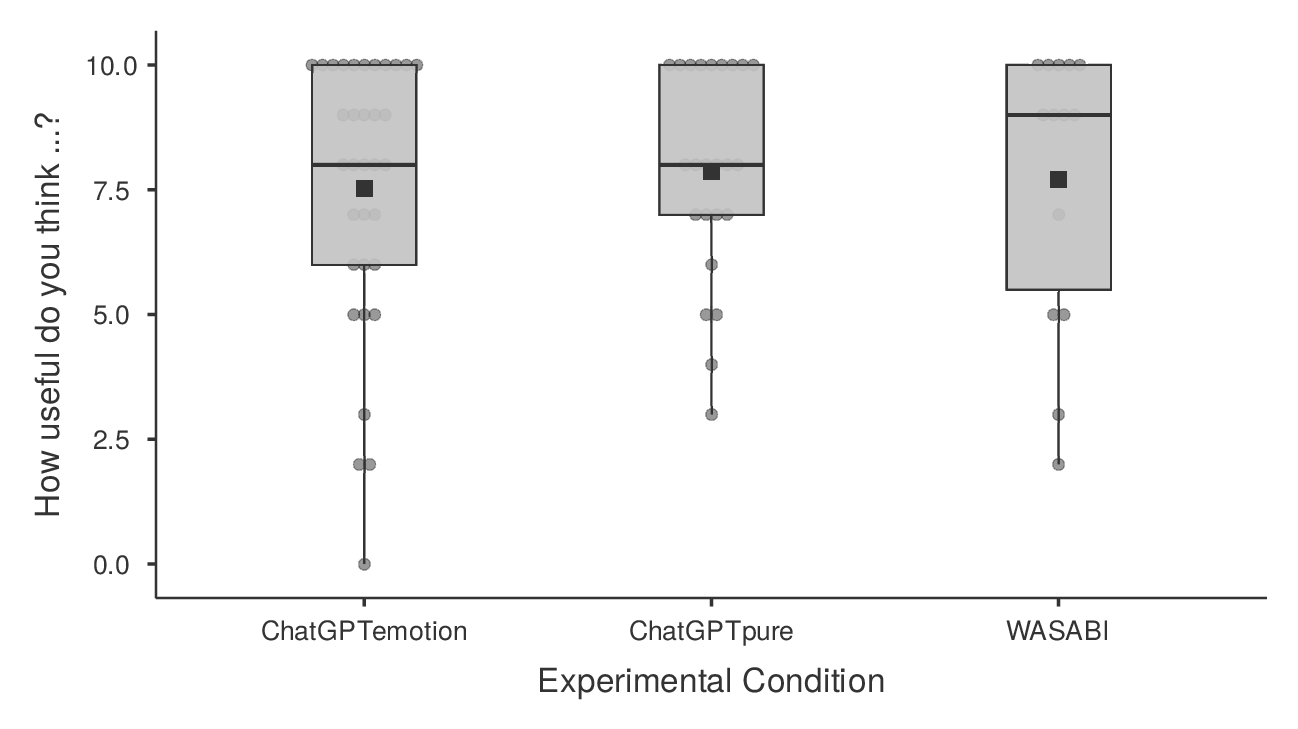}
    \caption{Box plot showing the distribution of the \textit{\acf{UT}} by experimental condition with indicated median values on a scale from zero to ten.}
    \Description{Box plots visualizing the descriptive data also reported in Table 1.}
    \label{usefulness_today_loc}
\end{minipage}
\end{figure*}

An overview of the remaining questionnaire results is presented in Figures~\ref{intention_to_use_loc}-\ref{usefulness_today_loc}. These box plots show that those visitors, who had interacted with the non-emotional version of the android robot (\textit{ChatGPTpure}), found it most useful (Figure~\ref{perceived_usefulness_loc}), easiest to use (Figure~\ref{perceived_ease_of_use_loc}) and had the highest intention to use it (Figure~\ref{intention_to_use_loc}). They also did not find the robot less useful today as compared to the groups of visitors who had interacted with one of the emotional variants (Figure~\ref{usefulness_today_loc}). 

To test these results for significance\footnote{All tests were performed using the \textit{JAMOVI} open source software (version 2.7.24.0).}, normality of the data was evaluated using the Shapiro–Wilk test for each variable and experimental condition, cf.~Table~\ref{tbl:Table_Descriptives_global}. For several groups the test indicated significant deviations from a normal distribution ($p < 0.05$, for \ac{ITU}, \ac{PU} and \ac{UT}). As the assumption of normality was not fully met in all dimensions, a non-parametric Kruskal–Wallis test was applied to compare the experimental conditions.

\begin{table}[!htbp]
\caption{Kruskal-Wallis test results}
\label{tbl:Table_Kruskal-Wallis_tam}
\centering
\begin{tabular}{lrrrr}
\toprule
~                            & $\chi^{2}$ & $df$ &    $p$ &   $\varepsilon^2$ \\
\midrule
IntentionToUse               &       3.55 &  2 & .169 & 0.05 \\
PerceivedUsefulness          &       5.76 &  2 & .056 & 0.08 \\
PerceivedEaseOfUse           &       6.83 &  2 & .033 & 0.09 \\
How useful do you think ...? &       0.13 &  2 & .939 & 0.00 \\
\bottomrule
\end{tabular}
\end{table}

The obtained $p$-values presented in Table~\ref{tbl:Table_Kruskal-Wallis_tam} for \ac{PU} and \ac{PEOU} remain close to or even below the desired level of significance ($\alpha = 0.05$) with a moderate effect size ($\varepsilon > 0.6$). Accordingly, pairwise comparisons were applied using the Dwass-Steel-Critchlow-Flinger test separately for these two dimensions. The results reveal that in both cases only the difference between \textit{ChatGPTemotion} and \textit{ChatGPTpure} is significant, cf.~Table~\ref{tbl:Table_Paarweise_Vergleiche_-_PerceivedUsefulness} and Table~\ref{tbl:Table_Paarweise_Vergleiche_-_PerceivedEaseOfUse}.

\begin{table}[!htbp]
\caption{Pairwise comparisons - PerceivedUsefulness}
\label{tbl:Table_Paarweise_Vergleiche_-_PerceivedUsefulness}
\centering
\begin{tabular}{llrr}
\toprule
~              & ~           &     $W$ &    $p$ \\
\midrule
ChatGPTemotion & ChatGPTpure &  3.39 & .044 \\
ChatGPTemotion & WASABI      & -0.14 & .995 \\
ChatGPTpure    & WASABI      & -2.11 & .296 \\
\bottomrule
\end{tabular}
\end{table}

\begin{table}[!htbp]
\caption{Pairwise comparisons - PerceivedEaseOfUse}
\label{tbl:Table_Paarweise_Vergleiche_-_PerceivedEaseOfUse}
\centering
\begin{tabular}{llrr}
\toprule
~              & ~           &     $W$ &    $p$ \\
\midrule
ChatGPTemotion & ChatGPTpure &  3.63 & .028 \\
ChatGPTemotion & WASABI      &  2.01 & .329 \\
ChatGPTpure    & WASABI      & -0.77 & .848 \\
\bottomrule
\end{tabular}
\end{table}


\begin{table}[!htbp]
\caption{Descriptives of the three questions regarding the emotionality of the robot: (1)  I have the impression that the robot displays emotional reactions, (2) I find the robot’s emotional responses appropriate to the situation, (3) The robot appears moody or shows emotional mood swings.}
\label{tbl:Table_Descriptives_emotion}
\centering
\begin{tabular}{llrrrrrr}
\toprule
Q.                                               & Condition &     $N$ & $Mean$ & $Median$ &   $SD$ \\
\midrule
(1) & ChatGPTemotion         & 35 & 4.40 &   4.00 & 1.83  \\
(1) & ChatGPTpure            & 24 & 4.04 &   4.00 & 1.99  \\
(1) & WASABI                 & 14 & 3.50 &   3.00 & 1.87  \\
(2)           & ChatGPTemotion         & 35 & 4.31 &   4.00 & 1.75  \\
(2)         & ChatGPTpure            & 24 & 4.54 &   4.50 & 2.11  \\
(2)         & WASABI                 & 14 & 4.43 &   4.00 & 1.99  \\
(3)                   & ChatGPTemotion         & 35 & 3.31 &   3.00 & 2.01  \\
(3)                   & ChatGPTpure            & 24 & 2.92 &   2.50 & 1.74  \\
(3)                   & WASABI                 & 14 & 3.00 &   3.00 & 1.92  \\
\bottomrule
\end{tabular}
\end{table}

The manipulation check whether the different visitors perceived the emotional expressions in the two conditions \textit{ChatGPTemotion} and \textit{WASABI} is presented in Table~\ref{tbl:Table_Descriptives_emotion}.
It can be seen that all averages lie close to four, which is the center point on the seven point likert scale. Also, there is reason to assume that the values are not distributed normally according to low $p$-values of the Shapiro-Wilk tests. The results of the Kruskal-Wallis test (cf.~Table~\ref{tbl:Table_Kruskal-Wallis_emo}) show no significant differences between conditions.

\begin{table}[!htbp]
\caption{Kruskal-Wallis test results for perceived emotionality / mood swings: (1) I have the impression that it displays emotional reactions, (2) I find its emotional responses appropriate to the situation, (3) The robot appears moody or shows emotional mood swings.}
\label{tbl:Table_Kruskal-Wallis_emo}
\centering
\begin{tabular}{lrrr}
\toprule
Q.  & $\chi^{2}$ & $df$ &    $p$ \\
\midrule
(1) &       2.59 &  2 & .274 \\
(2) &       0.29 &  2 & .866 \\
(3) &       0.51 &  2 & .773 \\
\bottomrule
\end{tabular}
\end{table}

\section{Discussion and conclusion}
\label{sec:discussion}

To the best of our knowledge, this work is the first to investigate the impact of emotional reactions of an interactive, autonomous, android robot in a public space. Unfortunately, the results seem to indicate that both of our  ways to implement an emotion process that derives emotions from user input and lets the robot express them via its face seem not to enhance its overall acceptance, but rather diminish it, although only to a very small extend. The negative result of the manipulation check further devalues our approach.
One possible reason for this may be the very public setup, which can keep people from engaging in deeply emotional conversations. In fact, inspecting the logs retrospectively revealed that emotions other than ``happy'' have hardly been displayed at all.

While the quantitative results of the survey remain somewhat inconclusive, qualitative observations made by the experimenters during the study provide valuable context regarding human-robot interaction in the wild. Notably, the robot's highly human-like appearance effectively camouflaged it within the museum environment. Many visitors initially mistook Andrea for a regular museum patron. The realization that it was an android often occurred only upon hearing it speak or after consulting the experimenters.

Furthermore, as noted in the study procedure, the initial threshold for engaging with the robot proved to be relatively high. The frequent need to actively encourage patrons and explain the microphone's usage highlights that spontaneous interaction with such systems is not yet fully intuitive. Once approached, visitors often required guidance on suitable conversation topics. Interactions typically commenced with cautious exploration, where users would ``test the waters'' by asking museum-related questions. However, as users became more accustomed to the system, the dialogue frequently expanded into an array of broader subjects, ranging from sports and the robot's technical capabilities to philosophical inquiries about the meaning of life. 

Language expectations also played a significant role in the visitors' initial hesitation, as many assumed the android was restricted to German or English. When the experimenters assured them that Andrea could comprehend and reply in their native languages, visitors exhibited visible satisfaction and a greater willingness to converse. This observation highlights the impact that accessible, multi-lingual capabilities have on fostering inclusive and engaging interactions in public spaces.

However, our study suffers from several limitations. With only $N=14$ questionnaire results the WASABI condition is likely underpowered and small effects remain undetectable. As explained, technical problems on the last day did not allow us to collect more data in this condition as we had planned initially. The robot's gesture behavior was rather limited as well. Apart from rising its left arm to its ear, making a fist with its left hand while looking down ``in thoughts'', and moving the arm back down again when starting to speak, Andrea did not use any iconic, metaphoric or even manipulative gestures.
Furthermore, the robot displays emotions only unimodally through facial expressions, while humans rely on additional modalities, like body posture and voice cues, to decode them \cite{keltner_understanding_2017}.

In our current work, first, we aim to improve the visibility of the emotional states by modulating the robot's speech according to the activated emotion at runtime dynamically. This can be achieved by deriving embedding vectors that represent emotions in the speaker embedding space of a \acl{TTS} system as, for example, proposed by the \textit{EmoKnob} \cite{chen2024emoknobenhancevoicecloning} framework.
Secondly, a personalized autobiographical memory has recently been implemented in the aim to let robot Andrea remember and reuse person-specific, conversational knowledge in its dialogue in long-term, repeated interactions. Extending the gesture behavior by integrating automated function calling in the LLM-pipeline is also an open task for future engineering work.

\begin{acks}
We would like to warmly thank the Mercedes-Benz Museum staff for their continuous support during the setup and the entire six-day exhibition, the students Leon Kiefer, Steve Aschenbrenner, and Dilara Celepci-Uludag for their help, as well as all museum visitors who kindly participated in our study and provided their feedback.
\end{acks}

\bibliographystyle{ACM-Reference-Format} 
\bibliography{references} 
\end{document}